\documentclass[10pt,twocolumn,letterpaper]{article}

\usepackage{cvpr}              

\usepackage{graphicx}
\usepackage{amsmath}
\usepackage{amssymb}
\usepackage{booktabs}
\usepackage{listings}
\usepackage{multirow}
\usepackage{xcolor}

\definecolor{codegreen}{rgb}{0,0.6,0}
\definecolor{codegray}{rgb}{0.5,0.5,0.5}
\definecolor{codepurple}{rgb}{0.58,0,0.82}
\definecolor{backcolour}{rgb}{0.95,0.95,0.92}

\lstdefinestyle{mystyle}{
  backgroundcolor=\color{backcolour}, commentstyle=\color{codegreen},
  keywordstyle=\color{magenta},
  numberstyle=\tiny\color{codegray},
  stringstyle=\color{codepurple},
  basicstyle=\ttfamily\footnotesize,
  breakatwhitespace=false,         
  breaklines=true,                 
  captionpos=b,                    
  keepspaces=true,                 
  numbers=left,                    
  numbersep=5pt,                  
  showspaces=false,                
  showstringspaces=false,
  showtabs=false,                  
  tabsize=2
}

\usepackage[pagebackref,breaklinks,colorlinks]{hyperref}

\usepackage[capitalize]{cleveref}
\crefname{section}{Sec.}{Secs.}
\Crefname{section}{Section}{Sections}
\Crefname{table}{Table}{Tables}
\crefname{table}{Tab.}{Tabs.}

\def\cvprPaperID{80} 
\def\confName{CVPR}
\def\confYear{2022}

\begin{document}

\title{Revisiting Vicinal Risk Minimization for \\Partially Supervised Multi-Label Classification \\
Under Data Scarcity
}

\author{Nanqing Dong\thanks{Corresponding Author}\\
Department of Computer Science\\
University of Oxford\\
{\tt\small nanqing.dong@cs.ox.ac.uk}
\and
Jiayi Wang\\
Mathematical Institute\\
University of Oxford
\and
Irina Voiculescu\\
Department of Computer Science\\
University of Oxford\\
{\tt\small irina.voiculescu@cs.ox.ac.uk}
}
\maketitle

\begin{abstract}
Due to the high human cost of annotation, it is non-trivial to curate a large-scale medical dataset that is fully labeled for all classes of interest. Instead, it would be convenient to collect multiple small partially labeled datasets from different matching sources, where the medical images may have only been annotated for a subset of classes of interest.
This paper offers an empirical understanding of an under-explored problem, namely partially supervised multi-label classification (PSMLC), where a multi-label classifier is trained with only partially labeled medical images. In contrast to the fully supervised counterpart, the partial supervision caused by medical data scarcity has non-trivial negative impacts on the model performance. A potential remedy could be augmenting the partial labels. Though vicinal risk minimization (VRM) has been a promising solution to improve the generalization ability of the model, its application to PSMLC remains an open question. To bridge the methodological gap, we provide the first VRM-based solution to PSMLC. The empirical results also provide insights into future research directions on partially supervised learning under data scarcity.
\end{abstract}

\section{Introduction}
\label{sec:intro}
Fueled by the joint development of theories~\cite{rumelhart1986learning,cybenko1989approximation,hornik1989multilayer}
and hardware, deep learning has led to a significant leap in computer-aided diagnosis, reaching or even outperforming human-level performance~\cite{rajpurkar2017chexnet}. As a data-driven method, DL models tend to require large-scale fully labeled images for supervised training. However, this is largely infeasible for many medical vision tasks due to high annotation costs, which gives rise to emerging research interests on partially supervised learning (PSL)~\cite{petit2018handling,gonzalez2018multi,durand2019learning,zhou2019prior,fang2020multi,shi2021marginal,xu2021partially,zhang2021dodnet,dong2022towards}.
Given a set of classes of interest, it is challenging to prepare a large dataset with all classes of interested annotated. Instead, it is more practical to source multiple relevant but partially labeled datasets, where each dataset is only annotated for a \emph{true} subset of classes of interests. This can be interpreted from the perspective of multi-task learning (MTL)~\cite{caruana1997multitask}, where the task of interest can be decomposed into multiple sub-tasks.

Existing PSL studies~\cite{petit2018handling,gonzalez2018multi,durand2019learning,zhou2019prior,fang2020multi,shi2021marginal,xu2021partially,zhang2021dodnet} tend to assume that large-scale partially labeled or even fully labeled data are available when designing the algorithms. However, this is infeasible in many specific scenarios, especially in the medical domain, where \emph{data scarcity} has been a topic of active research. Dong~\etal~\cite{dong2022towards} first proposed to use data augmentation to mitigate the data scarcity in partially supervised semantic segmentation, where vicinal risk minimization (VRM)~\cite{chapelle2001vicinal} is adopted to generate \emph{vicinal\/} fully labeled image-label pairs with only partially labeled data. Though how to address the data scarcity issue of partially supervised multi-label classification (PSMLC) remains an open question, inspired by \cite{dong2022towards}, we make a concrete first step towards it with VRM.

\begin{figure}[t]
\centering
\begin{subfigure}[t]{0.24\columnwidth}
    \includegraphics[width=\columnwidth]{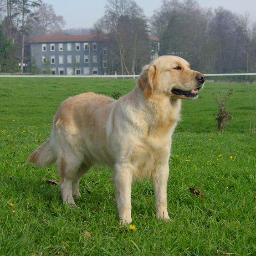}
    \caption{}
    \label{fig:obj1}
\end{subfigure}
\begin{subfigure}[t]{0.24\columnwidth}
    \includegraphics[width=\columnwidth]{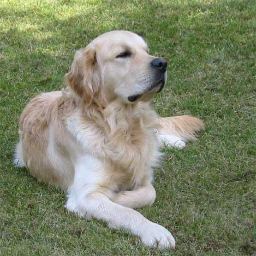}
    \caption{}
    \label{fig:obj2}
\end{subfigure}
\begin{subfigure}[t]{0.24\columnwidth}
    \includegraphics[width=\columnwidth]{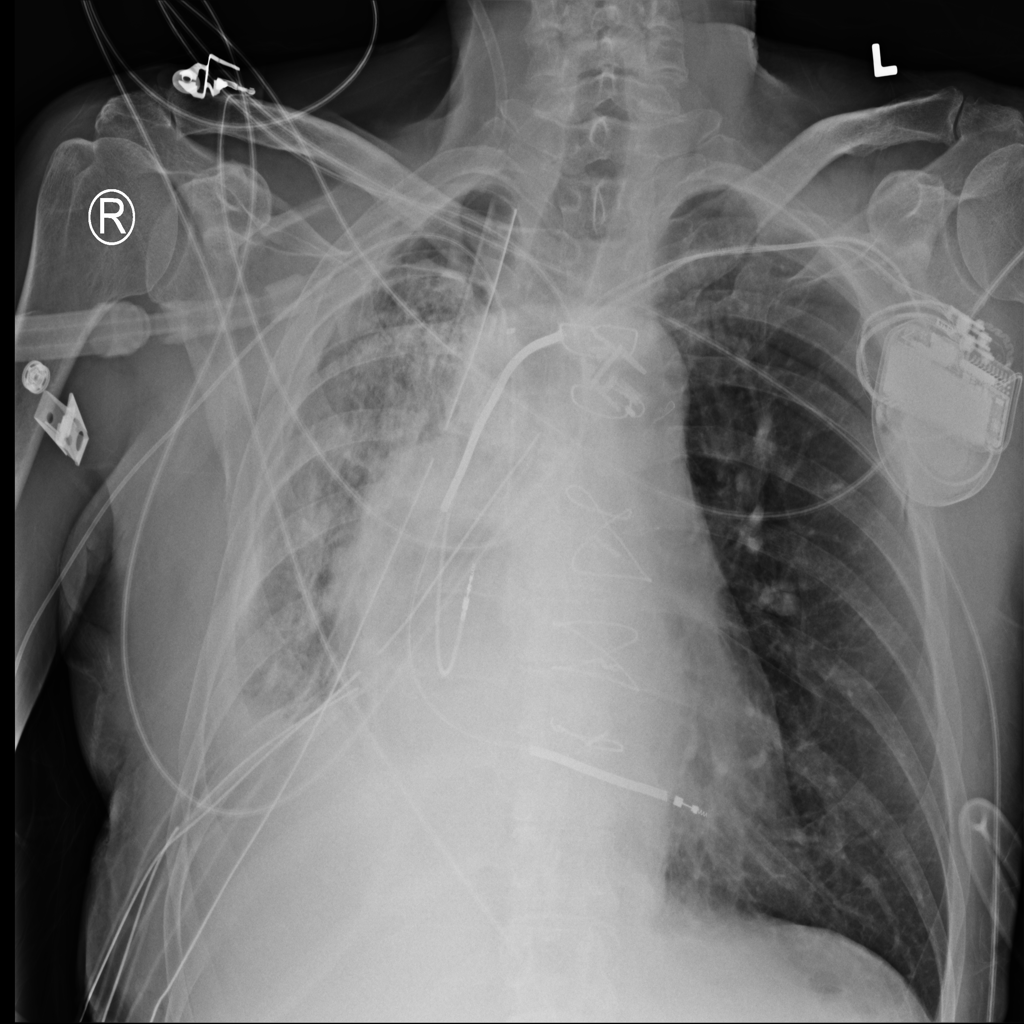}
    \caption{}
    \label{fig:obj3}
\end{subfigure}
\begin{subfigure}[t]{0.24\columnwidth}
    \includegraphics[width=\columnwidth]{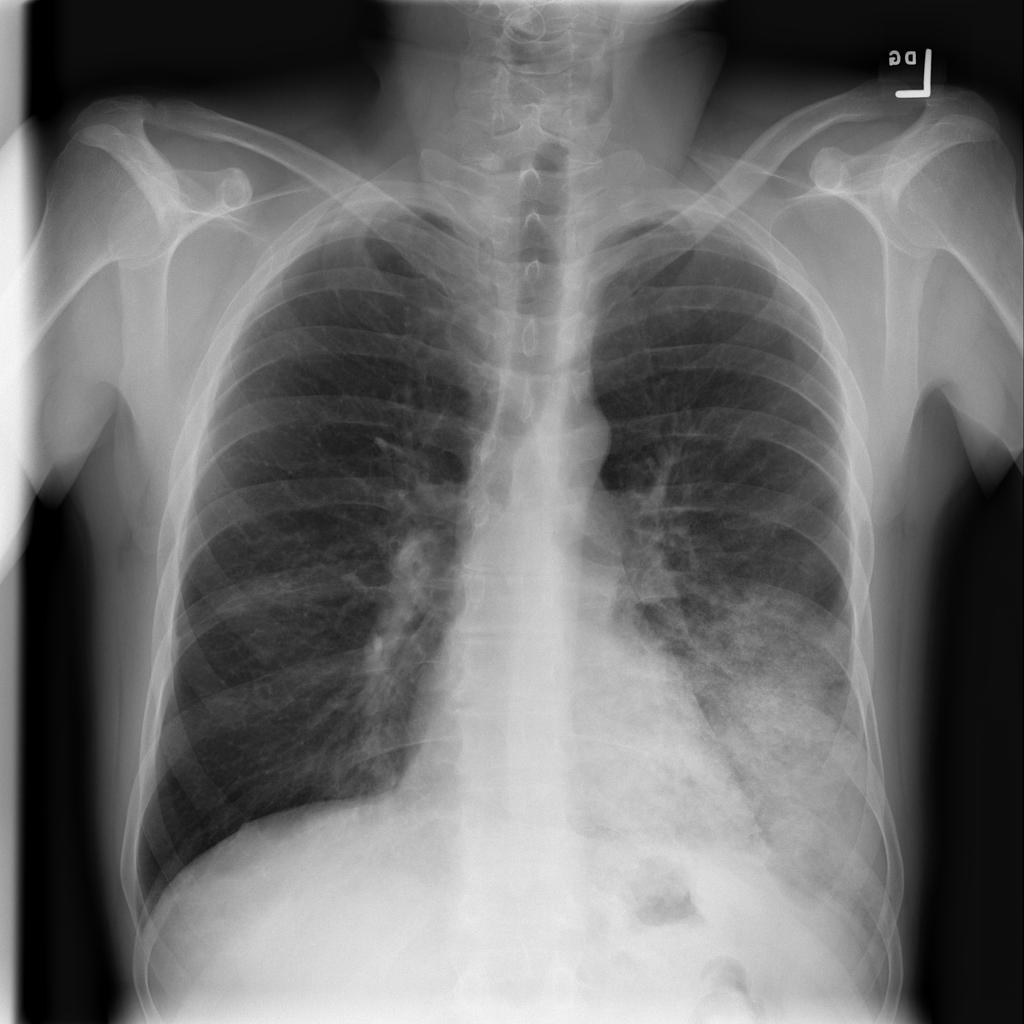}
    \caption{}
    \label{fig:obj4}
\end{subfigure}
\caption{(a) and (b) are two golden retrievers from ImageNet~\cite{deng2009imagenet}. (c) and (d) are two chest X-ray (CXR) images from ChestX-ray14~\cite{wang2017chestx}. In contrast to object-centric images, CXR images have multiple objects of semantic interests (\eg~organs). 
}
\label{fig:obj}
\end{figure}

\begin{figure*}[t]
\centering
\begin{subfigure}[t]{0.14\textwidth}
    \includegraphics[width=\columnwidth]{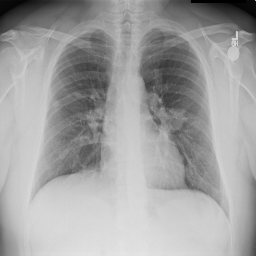}
    \includegraphics[width=\columnwidth]{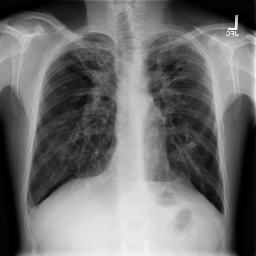}
    \caption{Input 1}
\end{subfigure}
\begin{subfigure}[t]{0.14\textwidth}
    \includegraphics[width=\columnwidth]{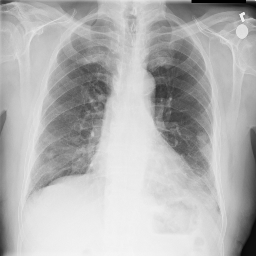}
    \includegraphics[width=\columnwidth]{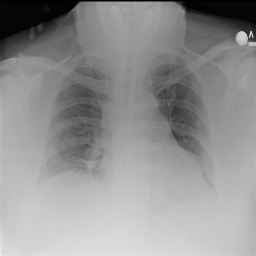}
    \caption{Input 2}
\end{subfigure}
\begin{subfigure}[t]{0.14\textwidth}
    \includegraphics[width=\columnwidth]{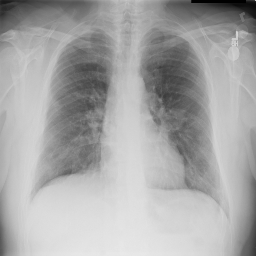}
    \includegraphics[width=\columnwidth]{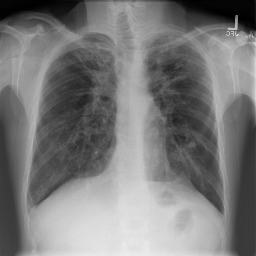}
    \caption{MixUp~\cite{zhang2018mixup}}
\end{subfigure}
\begin{subfigure}[t]{0.14\textwidth}
    \includegraphics[width=\columnwidth]{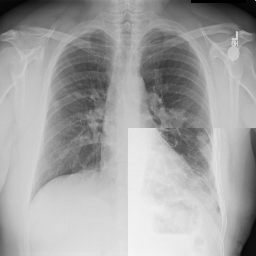}
    \includegraphics[width=\columnwidth]{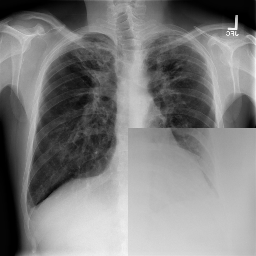}
    \caption{CutMix~\cite{yun2019cutmix}}
    \label{fig:vis:cutmix}
\end{subfigure}
\begin{subfigure}[t]{0.14\textwidth}
    \includegraphics[width=\columnwidth]{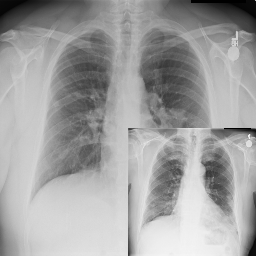}
    \includegraphics[width=\columnwidth]{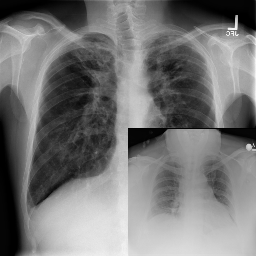}
    \caption{ResizeMix~\cite{qin2020resizemix}}
    \label{fig:vis:resizemix}
\end{subfigure}
\begin{subfigure}[t]{0.14\textwidth}
    \includegraphics[width=\columnwidth]{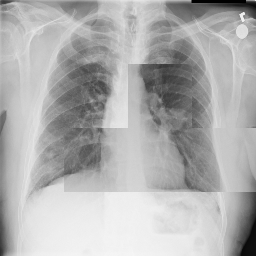}
    \includegraphics[width=\columnwidth]{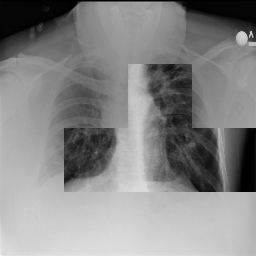}
    \caption{PuzzleMix~\cite{kim2020puzzle}}
    \label{fig:vis:puzzlemix}
\end{subfigure}
\caption{Illustration of state-of-the-art VRM methods on CXR images. First Row: Two CXR images are visually similar. Second Row: Two CXR images are visually different. The mixing ratio $\lambda$ is 0.75 for all four methods displayed. Firstly, compared with MixUp~\cite{zhang2018mixup}, CutMix~\cite{yun2019cutmix}, ResizeMix~\cite{qin2020resizemix}, and PuzzleMix~\cite{kim2020puzzle} generate less perceptually comfortable vicinal images (\ie~they do not look like real CXR images). Secondly, CutMix, ResizeMix, and PuzzleMix ignore the human structure similarity. Last but not least, in an MLC problem scenario, the vicinal images generated by CutMix, ResizeMix, and PuzzleMix might discard potential region of interests. For example, if the diseased region for the first input lies in the bottom right part of the CXR image, this might not be reflected in the vicinal images. In this work, we will focus on MixUp.
}
\label{fig:vis}
\end{figure*}

Various VRM-based data augmentation techniques~\cite{zhang2018mixup,yun2019cutmix,qin2020resizemix,kim2020puzzle,kim2021co,uddin2021saliencymix} have been designed to improve the generalization ability of a standard multi-class classifier trained on general object-centric images (\eg~\cref{fig:obj}). The first challenge is to generate vicinal images. Though many of these methods~\cite{yun2019cutmix,qin2020resizemix,kim2020puzzle} have reported state-of-the-art performance in multi-class classification tasks on general images, they tend to generate less meaningful vicinal images than MixUp~\cite{zhang2018mixup} in the medical domain. This phenomenon is due to the fact that these methods are designed for object-centric images where the mixing process can potentially keep the semantic information of interest. As illustrated in~\cref{fig:vis}, the vicinal images generated by CutMix~\cite{yun2019cutmix}, ResizeMix~\cite{qin2020resizemix}, and PuzzleMix~\cite{kim2020puzzle} can not utilize \emph{human structure similarity}~\cite{dong2018unsupervised,dong2021self,dong2022towards} or preserve the region of interests (\eg~the infectious regions) for medical images with multiple objects. As a comparison, MixUp shows robust visual performance over the other variants. Thus, we use MixUp to generate vicinal images in this work. The second challenge is to generate vicinal labels. The existing VRM methods are only designed for fully labeled data. Due to partial supervision, there are missing labels. That is to say, while we can randomly sample two images to generate a vicinal image with MixUp, we can not define the corresponding vicinal label with missing labels, which is deemed as the major bottleneck of applying VRM to PSMLC.

While previous efforts~\cite{durand2019learning} have been made to understand PSMLC with large-scale benchmark datasets from the perspective of label propagation~\cite{zhu2002learning}, the problem formulation of data scarcity in this study differentiates our contributions from~\cite{durand2019learning}. According to~\cite{dong2022towards}, VRM has shown unparalleled robustness in partially supervised semantic segmentation with only small-scale data. Motivated by this, we aim to leverage VRM to tackle PSMLC with data scarcity. Inspired by the principle of maximum entropy (PME)~\cite{jaynes1957information}, we propose a simple yet robust MixUp-based method for PSMLC, which can efficiently improve the model performance with only access to partial labels. We evaluate the proposed method via a set of controllable experiments on ChestX-ray14~\cite{wang2017chestx}, a public multi-label CXR dataset of thoracic conditions. In addition to providing initial empirical insights into PSMLC, we also validate that self-supervised pre-training~\cite{chen2021exploring} can further improve the model performance together with the proposed method. 

Our main contributions can be summarized as follows:
\begin{enumerate}
\item This is the first study of partially supervised multi-label classification (PSMLC) under data scarcity in the medical domain.
\item We adapt VRM to PSMLC and propose a simple yet robust PME-based technique to improve the model performance with only scarce partially labeled medical images.
\item The experimental results show initial empirical insights into future research directions on PSL under data scarcity. 
\end{enumerate}

\section{Related Work}
\paragraph{Partially Supervised Learning.}
Apart from~\cite{durand2019learning}, which provides an empirical understanding of PSMLC with large-scale general images (\eg~PASCAL VOC~\cite{everingham2010pascal} and MS COCO~\cite{lin2014microsoft},  the major breakthroughs in PSL lie in the field of multi-organ or multi-structure segmentation on medical images~\cite{petit2018handling,gonzalez2018multi,zhou2019prior,fang2020multi,shi2021marginal,xu2021partially,zhang2021dodnet,dong2022towards}. Gonzalez~\etal~\cite{gonzalez2018multi} first showed that, by only backpropagating the cross-entropy from the labeled part, the performance of multi-structure segmentation with sufficiently large-scale partially labeled data could be competitive with the performance of the fully supervised counterpart. However, in practice, it is unlikely to collect or curate a large amount of partially labeled data, \ie~only small-scale partially labeled datasets are available. Zhou~\etal~\cite{zhou2019prior} proposed to utilize a fully labeled set to learn image priors. However, the fully labeled set might be difficult to acquire in practice. Shi~\etal~\cite{shi2021marginal} proposed an \emph{exclusion loss} for mutually exclusive classes, which can not be used in MLC. Closely related to our work, Dong~\etal~\cite{dong2022towards} proposed to mitigate the data scarcity based on VRM. In contrast to this work, \cite{dong2022towards} requires that the classes of interest to be mutually exclusive, while our work is the first study on PSMLC.

\paragraph{Vicinal Risk Minimization.}
Zhang~\etal~\cite{zhang2018mixup} first proposed MixUp, a VRM-based data augmentation method in the input space by mixing two random images and the corresponding labels by convex interpolation. This idea has been further extended by many variants. For example, CutMix~\cite{yun2019cutmix} (\cref{fig:vis:cutmix}) replaces a random patch of the first image with a patch from the second image cropped at the same location.
ResizeMix~\cite{qin2020resizemix} (\cref{fig:vis:resizemix}) takes one step further by replacing a random patch of the first image with the resized second image. PuzzleMix~\cite{kim2020puzzle} (\cref{fig:vis:puzzlemix}) could be viewed as a generalization of CutMix, where the locations of the random patches are determined by saliency. As discussed in~\cref{sec:intro}, these variants of MixUp are not suitable for MLC or PSMLC tasks. 

\section{Preliminaries}
\paragraph{MixUp.} Let $(x_i, y_i)$ and $(x_j, y_j)$ be two image-label pairs randomly sampled from the training set $\mathcal{S} = \{(x_i, y_i)\}_{i=1}^n$. Here, $x \in \mathbb{R}^{H{\times}W{\times}C}$ is an image and $y \in  \mathbb{R}^{K}$ is considered as a one-hot encoded binary vector, where there are $K$ (mutually exclusive) classes of interest. The vicinal image-label pair $(\tilde{x}, \tilde{y})$ is defined as 
\begin{equation}
    \begin{split}
        \tilde{x} &= \lambda x_i + (1 - \lambda) x_j\\
        \tilde{y} &= \lambda y_i + (1 - \lambda) y_j\\
    \end{split}
    \label{eq:mixup}
\end{equation}
where $\lambda \sim \mathrm{Beta}(\alpha, \alpha)$ for $\alpha \in (0, \infty)$. From the perspective of MTL, a $K$-class MLC task could be decomposed into $K$ binary classification tasks, where MixUp can be applied.

\paragraph{Weighted Loss.} To combat class imbalance, the weighted binary cross-entropy (BCE) loss is commonly adopted in MLC. Given an image-label pair $(x, y)$, we use $y^i$ to denote the $i^{\mathrm{th}}$ entry of $y$. We have 
\begin{equation}
    \begin{split}
        \mathcal{L}(x, y^i) = &- w_{+} y^i \log p(y^i = 1 | x) \\
                               &- w_{-} (1 - y^i) \log p(y^i = 0 | x),
    \end{split}
    \label{eq:bce}
\end{equation}
where $w_{+} = \frac{n_{-}}{n_{+} + n_{-}}$ and  $w_{-} = \frac{n_{+}}{n_{+} + n_{-}}$ with $n_{+}$ and $n_{-}$ the number of positive and negative cases for the $i^{\mathrm{th}}$ class respectively.

\section{Partially Supervised Multi-Label Classification}
\subsection{Problem Formulation}
Without loss of generality, given a \emph{small-scale} training set $\mathcal{S} = \{(x_i, y_i)\}_{i=1}^n$, we define that, for each image-label pair $(x, y)$, at least one entry of $y$ is missing, \ie~$x$ is partially labeled, and for each class of interest, there are both positive and negative cases labeled. Given a model $f_\theta$, the goal of PSMLC is to maximize the prediction accuracy of $f_\theta$ with only \emph{limited} partially labeled data.

\subsection{MixUp with Partial Labels}
Now, we will explore how to adapt MixUp to PSMLC. For simplicity, we illustrate with an example of $K{=}2$. Again, $(x_i, y_i)$ and $(x_j, y_j)$ are two image-label pairs. where $y_i$ and $y_j$ are partial labels. There are two cases. 

\subsubsection{Locally Full Supervision}
\label{sec:method:local}
In the first case, $y_i$ and $y_j$ have partial labels for the same class, \eg~$y_i = [?, y_i^2]$ and $y_j = [?, y_j^2]$, where $?$ denotes the missing label for the $1^{\mathrm{st}}$ class and $y_i^2 \in \{0, 1\}, y_j^2 \in \{0, 1\}$. Obviously, \cref{eq:mixup} still holds if we only consider the $2^{\mathrm{nd}}$ class. Under the interpretation of MTL, we have \emph{locally} full supervision over the $2^{\mathrm{nd}}$ class. However, to leverage MixUp in this case, there will be additional computational cost in batch-wise sampling, which could be a non-trivial overhead in practice. A trivial solution is to decompose a PSMLC problem into multiple binary classification problems. However, when $K$ is large, this strategy will be inefficient as it does not utilize MTL.

\subsubsection{Globally Partial Supervision}
\label{sec:method:global}
We are more interested in the second case, where $y_i$ and $y_j$ have partial labels for different classes, \eg~$y_i = [?, y_i^2]$ and $y_j = [y_j^1, ?]$. In fact, in a partially labeled dataset, the majority of randomly sampled pairs will fall in this case, which is also the major bottleneck of PSMLC. Similar to~\cite{dong2022towards}, we aim to transform PSMLC into fully supervised MLC. However, \cite{dong2022towards} defines the \emph{vicinity distribution}~\cite{chapelle2001vicinal} by utilizing human structure similarity, which is infeasible for PSMLC. Instead, we regularize the vicinity distribution by a simple probability trick.

\begin{figure}[t]
    \centering
    \includegraphics[width=\columnwidth]{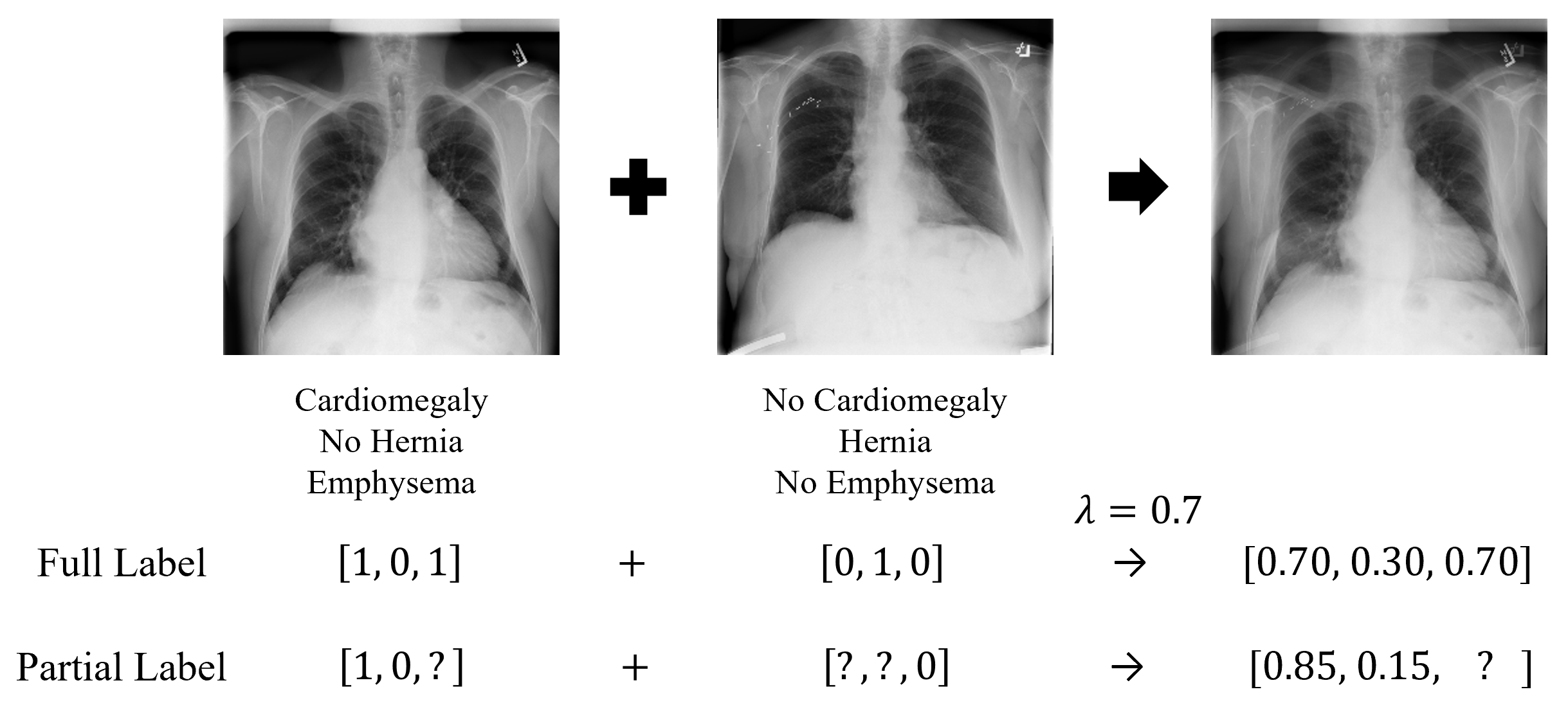}    
    \caption{Illustration of MixUp based on PME with two CXR images with 3 classes of interest. For full labels, \cref{eq:mixup} can be directly applied. For partial labels, \cref{eq:pme} is only applied for the classes where the first image has no missing labels.}
    \label{fig:demo}
\end{figure}

The principle of maximum entropy (PME)~\cite{jaynes1957information} was first proposed in 1950s. With limited prior knowledge over the unknown true distribution, PME defines the vicinity distribution with the largest entropy. Concretely, for the $k^{\mathrm{th}}$ class, assume only $x_i$ has the partial label (\eg~$y_i^k = 1$) and $x_j$ has missing label ($y_j^k \leftarrow ?$), without any prior knowledge about $x_j$, we define $\hat{y}_j^k = 0.5$, \ie~considering $y_j^k$ as an independent system with two possible states, discrete uniform distribution leads to the largest entropy. We define the vicinal label as
\begin{equation}
    \begin{split}
        \tilde{x} &= \lambda x_i + (1 - \lambda) x_j \\
        \tilde{y}^k &= \lambda y_i^k + (1 - \lambda) \hat{y}_j^k \\
    \end{split}
    \label{eq:pme}
\end{equation}
where $\lambda \sim \mathrm{Uniform}(\alpha_k, 1)$ and $0.5 \le \alpha_k < 1$ is a hyperparameter. Note, we use $\alpha_k$ instead of $\alpha$ as the choice of $\alpha_k$ can be dependent on the $k^{\mathrm{th}}$ class. This is different from MixUp-based methods~~\cite{zhang2018mixup,yun2019cutmix,qin2020resizemix,kim2020puzzle,kim2021co,uddin2021saliencymix} that are designed for multi-class classification tasks. We will give more details of this design in \cref{sec:exp:result}. Though~\cref{eq:pme} has a similar format to~\cref{eq:mixup}, we want to highlight a few difference: (a) $x_i$ should have locally full supervision for the $k^{\mathrm{th}}$ class (\eg~$y_i^k {=}1$); (b) $\lambda$ is required to be larger than 0.5 for the known example ($x_i$ in this case). As we are certain about $y_i^k$, there are only two possible states for $y_j^k$, \ie~$y_j^k{=}1$ or $y_j^k{=}0$. It can be inferred that if $y_i^k{=}1$, we have $0.75 < \tilde{y}^k = 0.5 + 0.5\lambda < 1$; if $y_i^k = 0$, we have $\tilde{y}^k = 0.5 (1{-}\lambda) < 0.25$. Clauses (a) and (b) ensure that the generated vicinal label $\tilde{y}^k$ has a logically reasonable label distribution. An illustrative comparison between MixUp and the proposed method is presented in \cref{fig:demo}.

\paragraph{Relationship with Noisy Labels} Learning from noisy labels~\cite{natarajan2013learning} has been a seminal strategy in utilizing unlabeled data over the past decade~\cite{zhu2002learning}. Here, we define the noisy labels as the pseudo labels that are automatically generated by the algorithms for the unlabeled data. This strategy has shown state-of-the-art performance in semi-supervised learning benchmark tasks~\cite{sohn2020fixmatch}. The problem of interest shares a similar problem formulation with semi-supervised learning, where unlabeled data exist. However, the proposed method differs from semi-supervised learning in three aspects. First, instead of learning the pseudo labels, we generate the pseudo labels by PME. That means, the proposed method is computationally efficient. Second, essentially, the proposed method is a data augmentation technique, which simultaneously augments the base image ($x_i$ in \cref{eq:pme}) and adds noise to the base label ($y_i^k$ in \cref{eq:pme}). In fact, adding noise to the ground truth labels has been shown as an effective method to improve the model robustness and generalization ability~\cite{vahdat2017toward}. Last but not least, the generated pseudo labels ($\tilde{y}^k$) are logically reliable. As shown in~\cite{dong2022towards}, semi-supervised learning-based noisy labels might not be reliable if there are not enough labeled data to learn the model. In the problem of formulation, we focus on a data scarcity situation. Thus, a semi-supervised learning approach is not feasible.

\subsection{Training Strategy}
\label{sec:method:train}
For fully labeled images, the linear combination in~\cref{eq:mixup} can be performed efficiently with batch-wise processing, as shown in~\cite{zhang2018mixup}. However, it requires additional computational cost to sample two image-label pairs. First, we have to decompose the task into $K$ binary classification sub-tasks, which gives up the formulation of MTL and increases the number of forward and backward passes in the optimization process. Second, when sampling a pair, we need to make sure that two images are partially labeled for the same class of interest. Third, when $K$ is large, two image-label pairs can have both locally full supervision and globally partial supervision. To fully utilize the advantages of random sampling in stochastic gradient descent, we need to efficiently implement \cref{eq:mixup} and \cref{eq:pme} simultaneously in the batch processing.

To solve the above issues, we present a computation-efficient implementation for batch-wise training with only 8 lines of PyTorch code, shown in~\cref{code}.\footnote{For simplicity, we use the same value of $\texttt{alpha}$ for different classes to illustrate the main concept of the proposed method.}

\begin{lstlisting}[language=Python, label=code, caption=Batch-wise training in PyTorch.]
# x1, x2: batch of images, [B, 3, H, W]
# y1, y2: batch partial labels where the missing 
#         entries are filled with 0.5, [B, K]
# alpha: hyperparameter
# model: neural network
# criterion: loss function with reduction='none'
lam = numpy.random.uniform(alpha, 1)
mask = y1 != 0.5
x = lam * x1 + (1. - lam) * x2
y = lam * y1 + (1. - lam) * y2
loss = (criterion(model(x), y) * mask).mean()
optimizer.zero_grad()
loss.backward()
optimizer.step()
\end{lstlisting}

We also provide a concrete example for a better illustration of the mechanism behind our implementation. Let $y_1 = [1, ?, 0, ?]$ and $y_2 = [1, 1, ?, ?]$ be two label vectors with missing labels (\ie~$K = 4$). Given an arbitrary $\lambda \sim \mathrm{Uniform}(0.5, 1)$ (say $\lambda = 0.75$), a step-by-step demonstration of~\cref{code} is provided in~\cref{tab:demo}.

\begin{table}[h]
    \centering
    \begin{tabular}{|c | c | c | c | }
    \hline
    Step & $y_1$ & $y_2$ & mask\\\hline
    -- & [1, ?, 0, ?] & [1, 1, ?, ?] & -- \\\hline
    Fill $?$  & \multirow{2}{*}{[1, 0.5, 0, 0.5]} & \multirow{2}{*}{[1, 1, 0.5, 0.5]} & \multirow{2}{*}{--} \\
    with 0.5 & & & \\\hline
    Get  & \multirow{2}{*}{[1, 0.5, 0, 0.5]} & \multirow{2}{*}{[1, 1, 0.5, 0.5]} & \multirow{2}{*}{[1, 0, 1, 0]} \\
    mask& & & \\\hline
    Get  & \multicolumn{2}{c|}{\multirow{2}{*}{[1, 0.625, 0.25, 0.5]}} & \multirow{2}{*}{[1, 0, 1, 0]} \\
    $\tilde{y}$ & \multicolumn{2}{c|}{} & \\\hline
    \end{tabular}
    \caption{A step-by-step demonstration of \cref{code}. After getting $\tilde{y}$ with \cref{eq:mixup} (or \cref{eq:pme}) with $\lambda = 0.75$, the BCE losses for $K=4$ classes are computed following~\cref{eq:bce}. However, only the BCE losses of the first and the third classes will be back-propagated as the BCE losses of the second and the fourth classes are zeroed out.}
    \label{tab:demo}
\end{table}

\section{Experiments}
\subsection{Experimental Setup}

\paragraph{Baselines.} We consider four models in our experiments, where the four models share the same network backbone. The first one is a standard MLC model where the missing labels are ignored in the backpropagation~\cite{gonzalez2018multi}. We denote this model as \emph{vanilla}. For the second model, we apply MixUp in a locally full supervision fashion to train an MLC model, where \cref{eq:mixup} is applied as described in~\cref{sec:method:local}. Following~\cite{zhang2018mixup}, we set $\alpha = 1$. We use the default valThe second baseline is denoted as MixUp. The third model is the proposed PME-based MixUp variant. Note, the third model is a unification of locally full supervision and globally partial supervision, where \cref{eq:mixup} and \cref{eq:pme} are efficiently integrated (as shown in \cref{sec:method:train}). For simplicity, we denote the proposed method as MixUp-PME. To understand the impact of PSMLC under data scarcity, we present the fourth model, which is the same MLC model trained with full labels. We denote this model as \emph{Oracle}.

\paragraph{Data.} We use the ChestX-ray14~\cite{wang2017chestx} public multi-label dataset of thoracic conditions,\footnote{\url{https://nihcc.app.box.com/v/ChestXray-NIHCC}} and adopt its default batch splits to ensure reproducibility. We use the first 1000 CXR images of the first batch as the training set and the second 1000 CXR images of the first batch as the test set. For simplicity, we illustrate the proposed method with a simple case that each image is only labeled for one class. We generate the partially labeled datasets by choosing 4 most common diseases among 14 identified conditions (\ie~$K = 4$), which are \emph{infiltration}, \emph{effusion}, \emph{atelectasis}, and \emph{nodule}.

\paragraph{Implementation.} Following the setup of \cite{rajpurkar2017chexnet}, we use DenseNet121~\cite{huang2017densely} as the network backbone. We minimize the weighted loss (\cref{eq:bce}) by using a standard Adam optimizer~\cite{kingma2015adam} with fixed learning rate $10^{-3}$ and batch size 64. In the inference phase, we use 0.5 as the default threshold for the predicted probability score. We train all the baselines for 30 epochs and report the best mean F1-score for each baseline, where
\begin{equation}
    F_1 = 2~\frac{precision * recall}{precision + recall},
    \label{eq:f1}
\end{equation}
\ie~the harmonic mean of precision and recall. We run each experiments for three times with different random seeds. All experiments are conducted in PyTorch~\cite{paszke2019pytorch} on an NVIDIA Tesla V100. Note, the traditional data augmentation that manipulates the image space does not directly solve the partial supervision problem~\cite{dong2022towards}. Thus, for a fair comparison, we  do not involve any traditional data augmentation. All CXR images are resized to a fixed size of $224{\times}224$. As a pre-processing step, instance normalization~\cite{dai2018scan} is performed on each CXR image:
\begin{equation}
    \hat{x}^{ij} =  \frac{x^{ij} - \mu(x)}{\sigma(x)},
\end{equation}
where $x$ is an image, $\hat{x}$ is the normalized image, $(i,j)$ is the position of the pixel, and $\mu$ and $\sigma$ are the mean and standard deviation of the pixels of $x$.

\begin{figure}[t]
    \centering
    \includegraphics[width=\columnwidth]{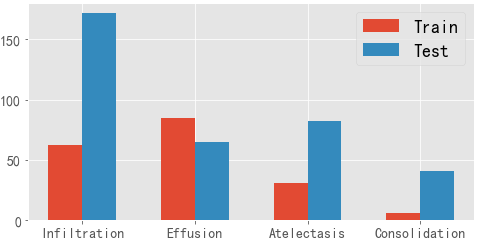}    
    \caption{Label statistics of positive cases for the partially labeled training set and fully labeled test set.}
    \label{fig:label}
\end{figure}

\begin{table*}[t]
    \centering
    \begin{tabular}{|c|c|c|c|c|c|}\hline
    Model & \emph{Infiltration} & \emph{Effusion} & \emph{Atelectasis} & \emph{Nodule} & Average \\\hline
    \emph{vanilla} & 0.3031 & 0.1807 & 0.1807 & 0.0704 & 0.1837 \\\hline
    MixUp & 0.2959 & 0.1701 & 0.1682 & 0.0707 & 0.1762 \\\hline
    MixUp-PME & 0.2949 & 0.1327 & 0.1584 & \textbf{0.2151} & \textbf{0.2002} \\\hline
    AMP & 0.3034 & \textbf{0.1894} & \textbf{0.1889} & 0.0788 & 0.1901\\\hline\hline
    \emph{Oracle} & \textbf{0.3547} & 0.1710 & 0.1739 & 0.0699 & 0.1924 \\\hline
    \end{tabular}
    \caption{Performance comparison for PSMLC. In the training set, each CXR image is only partially labeled for one class.}
    \label{tab:mlc}
\end{table*} 

\begin{figure}[t]
    \centering
    \includegraphics[width=\columnwidth]{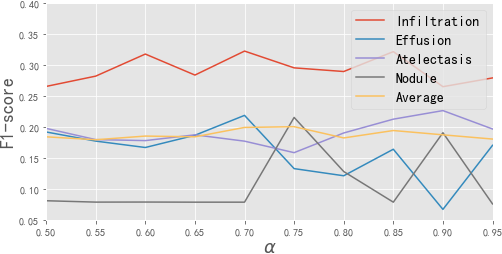}    
    \caption{Sensitivity of MixUp-PME to $\alpha$ for different classes.}
    \label{fig:alpha}
\end{figure}

\subsection{Empirical Analysis}
\label{sec:exp:result}
To provide a comprehensive understanding of the problem of interest, we consider a simple situation in the first experiment. The training set is equally split into 4 subsets, where each subset only contains labels for one class. The label distributions of the partially labeled training set and the fully labeled test set are summarized in \cref{fig:label}. In this experiment, we use the same value of $\alpha$ for different classes. The numerical results for four models are presented in \cref{tab:mlc}, where we report the mean F1-score for four classes. For MixUp-PME, we report the performance of $\alpha = 0.75$ in \cref{tab:mlc}, which gives the highest mean F1-score. The complete results of MixUp-PME under different values of $\alpha$ are depicted in \cref{fig:alpha}. Based on \cref{tab:mlc} and \cref{fig:alpha}, there are four empirical findings. First, an MLC model trained with full labels might not outperform the one trained with partial labels. Second, MixUp might not improve the performance of PSMLC. Third, MixUp-PME can significantly improve the performance of the class(es) under extreme class imbalance. Fourth, the performance of MixUp-PME is sensitive to the value of $\alpha$.

\paragraph{Adaptive MixUp-PME.} We notice an interesting phenomenon in \cref{fig:alpha}: while effusion achieves higher performance with smaller $\alpha$, atelectasis and nodule tend to achieve higher performance with larger $\alpha$. Intuitively, a MLC task can be decomposed into $K$ different sub-tasks. It can be seen in \cref{fig:alpha} that the highest F1-score that can be achieved by MixUp-PME for each class is higher than the counterpart achieved by \emph{vanilla}.\footnote{The highest F1-score achieved by MixUp-PME under different values of $\alpha$ \vs the F1-score achieved by \emph{vanilla}: 0.3219 \vs 0.3031 (infiltration), 0.2184 \vs 0.1807 (effusion), 0.2261 \vs 0.1807 (atelectasis), 0.2002 \vs 0.0704 (nodule).} This means that, depending on the class imbalance situation and difficulty of the sub-task, $\alpha$ could be adaptive to different classes to improve the overall MTL performance. Thus, a reasonable hypothesis is \textit{each sub-task should have an independent $\alpha$}. To validate this hypothesis, we repeat the experiment of MixUp-PME with the suitable values of $\{\alpha_k\}_{k=1}^K$ inferred from \cref{fig:alpha}. We denote this adaptive design as Adaptive MixUp-PME (AMP). The results are shown in \cref{tab:mlc}, where AMP outperforms \emph{vanilla} on all four classes this time. Surprisingly, AMP even outperforms \emph{Oracle} on three classes. It is worth mentioning that, in this work, we use the mean F1-score as the major performance measurement. In practice, AMP might be preferred than MixUp-PME if the performance of individual class is more important than the average performance.

\begin{figure}[t]
    \centering
    \includegraphics[width=\columnwidth]{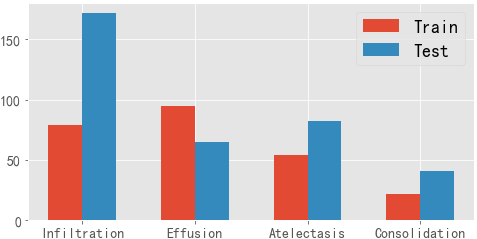}    
    \caption{Label statistics of positive cases for the partially labeled training set and fully labeled test set.}
    \label{fig:label_random}
\end{figure}

\begin{table*}[t]
    \centering
    \begin{tabular}{|c|c|c|c|c|c|}\hline
    Model & \emph{Infiltration} & \emph{Effusion} & \emph{Atelectasis} & \emph{Nodule} & Average \\\hline
    \emph{vanilla} & 0.3096 & 0.1341 & 0.1514 & 0.0553 & 0.1626 \\\hline
    MixUp-PME & 0.3139 & \textbf{0.1717} & \textbf{0.2024} & \textbf{0.0783} & \textbf{0.1916} \\\hline\hline
    \emph{Oracle} & \textbf{0.3547} & 0.1710 & 0.1739 & 0.0699 & 0.1924 \\\hline
    \end{tabular}
    \caption{Performance comparison for PSMLC. In the training set, each CXR image could be labeled for multiple classes. MixUp can not be applied under this situation.}
    \label{tab:random}
\end{table*}

\begin{table*}[t]
    \centering
    \begin{tabular}{|c|c|c|c|c|c|c|}\hline
    \multicolumn{2}{|c|}{} & \emph{Infiltration} & \emph{Effusion} & \emph{Atelectasis} & \emph{Nodule} & Average \\\hline
    \multirow{3}{*}{Exp 1} & \emph{vanilla} & 0.3031 & 0.1807 & 0.1807 & 0.0704 & 0.1837 \\\cline{2-7}
    & \emph{w/o} SSL & 0.2949 & 0.1327 & 0.1584 & \textbf{0.2151} & \textbf{0.2002} \\\cline{2-7}
     & \emph{w/} SSL & \textbf{0.3293} &  \textbf{0.1953} & \textbf{0.1948} &  0.0772 & 0.1991 \\\hline\hline
    \multirow{3}{*}{Exp 2} & \emph{vanilla} & 0.3096 & 0.1341 & 0.1514 & 0.0553 & 0.1626 \\\cline{2-7}
    & \emph{w/o} SSL & 0.3139 & \textbf{0.1717} & 0.2024 & 0.0783 & 0.1916 \\\cline{2-7}
     & \emph{w/} SSL & \textbf{0.3478} & 0.1277 & \textbf{0.2141} & \textbf{0.0912} & \textbf{0.1952}\\\hline\hline
    \multicolumn{2}{|c|}{\emph{Oracle}} & \textbf{0.3547} & 0.1710 & 0.1739 & 0.0699 & 0.1924 \\\hline
    \end{tabular}
    \caption{Impact of self-supervised pre-training on MixUp-PME. ``\emph{w/o} SSL`` denotes that the model is not pre-trained. ``\emph{w/} SSL`` denotes that the model is pre-trained.}
    \label{tab:ssl}
\end{table*} 

\paragraph{Robustness under MTL} One limitation of the experiment in \cref{tab:mlc} is that only single class is partially labeled for each image. Under the setup of the first experiment, MixUp-PME might not be able to fully leverage the advantage of MTL, which is one of advantage of the proposed method. In the second experiment, we consider a challenging situation that each CXR image can be labeled for more than one class. To simulate this situation, for each class of each image, we randomly generate a binary number (0 or 1) from a Bernoulli distribution with equal possibilities (\ie~$\textrm{Bernoulli}(0.5)$). The label distributions of the simulated partially labeled training set and the fully labeled test set are summarized in \cref{fig:label_random}. Note, in the experiment, MixUp can not be applied anymore as it is difficult to find two image-label pairs with the same set of labeled classes. Moreover, under this simulation, CXR images could also be unlabeled or fully labeled. We set $\alpha = 0.75$ for MixUp-PME following the first experiment. The quantitative comparison between MixUp-PME and \emph{vanilla} is presented in \cref{tab:random}. Compared with \cref{tab:mlc}, the performance of \emph{vanilla} is negatively influenced by this challenging experimental setup. On the contrary, MixUp-PME benefits from MTL with a huge performance gain. MixUp shows robust performance by outperforming \emph{vanilla} on all four classes by a large margin and outperforming \emph{Oracle} on three classes.

\paragraph{Impact of Unsupervised Pre-Training.} Learning transferable representations from unlabeled data then fine-tuning with limited labels has been shown as a label-efficient learning paradigm. We leverage a state-of-the-art self-supervised learning (SSL) framework SimSiam~\cite{chen2021exploring} to pre-train the network backbone and repeat the first experiment above. Here, we assume additional large-scale unlabeled data are available. The pre-training is performed on 4 batches of ChestX-ray14 for 200 epochs.\footnote{We use the second, the third, the fourth, and the fifth batches as the pre-training dataset, which contains $10^4$ CXR images in total.} We repeat the first and the second experiments, while this time, the models are initialized with the pre-trained weights. The results are shown in \cref{tab:ssl}. With self-supervised pre-training, MixUp-PME further improves its performance over several classes and outperforms \emph{vanilla} by a large margin. We conclude that SSL can be utilized to boost model performance under data scarcity.

\section{Limitations and Future Directions}
\paragraph{Data Decentralization.} A fundamental assumption of this study is that the training partially labeled datasets can be collected and stored in a centralized fashion. However, in practice, especially in the medical domain, the decentralized datasets are stored in different hospitals~\cite{dong2021federated}. Under the data regulations, it is not possible to apply MixUp-PME or AMP without exchanging users' data. An emerging research direction is federated PSMLC.

\paragraph{Domain Shift.} In the experiments, we only consider the data scarcity and class imbalance. In addition, \emph{domain shift}~\cite{ben2007analysis} could be a practical problem when collecting datasets from different sources~\cite{dong2018unsupervised}. The discussion on domain shift is left for future work.

\paragraph{Hyperparameters for AMP.} In our second experiment, we choose the set of $\{\alpha_k\}_{k=1}^K$ based on posterior knowledge on the first experiment. In practice, a similar trick can be applied to find the suitable values of $\{\alpha_k\}_{k=1}^K$ on a small validation set. However, when $K$ is large, this process could be troublesome. There is a trade-off between the performance and computational cost when applying AMP.

\section{Conclusion}
We present the first study of partially supervised multi-label classification (PSMLC)  under data scarcity, an unexplored but practical problem in PSL. We propose a novel VRM method that is based the principle of maximum entropy. The experimental results show that the proposed method can be used to mitigate the data scarcity issue of PSMLC. In the future work, we will explore PSMLC under more data challenges.

\paragraph{Acknowledgements.}
The authors would like to thank Huawei Technologies Co., Ltd. for providing GPU computing service for this study.
\clearpage
{\small
\bibliographystyle{ieee_fullname}
\bibliography{refs}

\begin{thebibliography}{10}\itemsep=-1pt

\bibitem{ben2007analysis}
Shai Ben-David, John Blitzer, Koby Crammer, and Fernando Pereira.
\newblock Analysis of representations for domain adaptation.
\newblock In {\em NIPS}, pages 137--144, 2007.

\bibitem{caruana1997multitask}
Rich Caruana.
\newblock Multitask learning.
\newblock {\em Machine Learning}, 28(1):41--75, 1997.

\bibitem{chapelle2001vicinal}
Olivier Chapelle, Jason Weston, L{\'e}on Bottou, and Vladimir Vapnik.
\newblock Vicinal risk minimization.
\newblock In {\em NIPS}, pages 416--422, 2001.

\bibitem{chen2021exploring}
Xinlei Chen and Kaiming He.
\newblock Exploring simple siamese representation learning.
\newblock In {\em CVPR}, pages 15750--15758, 2021.

\bibitem{cybenko1989approximation}
George Cybenko.
\newblock Approximation by superpositions of a sigmoidal function.
\newblock {\em Mathematics of control, signals and systems}, 2(4):303--314,
  1989.

\bibitem{dai2018scan}
Wei Dai, Nanqing Dong, Zeya Wang, Xiaodan Liang, Hao Zhang, and Eric~P Xing.
\newblock Scan: Structure correcting adversarial network for organ segmentation
  in chest x-rays.
\newblock In {\em Deep Learning in Medical Image Analysis and Multimodal
  Learning for Clinical Decision Support}, pages 263--273. Springer, 2018.

\bibitem{deng2009imagenet}
Jia Deng, Wei Dong, Richard Socher, Li-Jia Li, Kai Li, and Li Fei-Fei.
\newblock Imagenet: A large-scale hierarchical image database.
\newblock In {\em CVPR}, pages 248--255. IEEE, 2009.

\bibitem{dong2018unsupervised}
Nanqing Dong, Michael Kampffmeyer, Xiaodan Liang, Zeya Wang, Wei Dai, and Eric
  Xing.
\newblock Unsupervised domain adaptation for automatic estimation of
  cardiothoracic ratio.
\newblock In {\em MICCAI}, pages 544--552. Springer, 2018.

\bibitem{dong2022towards}
Nanqing Dong, Michael Kampffmeyer, Xiaodan Liang, Min Xu, Irina Voiculescu, and
  Eric Xing.
\newblock Towards robust partially supervised multi-structure medical image
  segmentation on small-scale data.
\newblock {\em Applied Soft Computing}, page 108074, 2022.

\bibitem{dong2021self}
Nanqing Dong, Michael Kampffmeyer, and Irina Voiculescu.
\newblock Self-supervised multi-task representation learning for sequential
  medical images.
\newblock In {\em ECML}, pages 779--794. Springer, 2021.

\bibitem{dong2021federated}
Nanqing Dong and Irina Voiculescu.
\newblock Federated contrastive learning for decentralized unlabeled medical
  images.
\newblock In {\em MICCAI}, pages 378--387. Springer, 2021.

\bibitem{durand2019learning}
Thibaut Durand, Nazanin Mehrasa, and Greg Mori.
\newblock Learning a deep convnet for multi-label classification with partial
  labels.
\newblock In {\em CVPR}, pages 647--657, 2019.

\bibitem{everingham2010pascal}
Mark Everingham, Luc Van~Gool, Christopher~KI Williams, John Winn, and Andrew
  Zisserman.
\newblock The pascal visual object classes (voc) challenge.
\newblock {\em IJCV}, 88(2):303--338, 2010.

\bibitem{fang2020multi}
Xi Fang and Pingkun Yan.
\newblock Multi-organ segmentation over partially labeled datasets with
  multi-scale feature abstraction.
\newblock {\em IEEE TMI}, 2020.

\bibitem{gonzalez2018multi}
Germ{\'a}n Gonz{\'a}lez, George~R Washko, and Ra{\'u}l San~Jos{\'e}
  Est{\'e}par.
\newblock Multi-structure segmentation from partially labeled datasets.
  application to body composition measurements on ct scans.
\newblock In {\em Image Analysis for Moving Organ, Breast, and Thoracic
  Images}, pages 215--224. Springer, 2018.

\bibitem{hornik1989multilayer}
Kurt Hornik, Maxwell Stinchcombe, and Halbert White.
\newblock Multilayer feedforward networks are universal approximators.
\newblock {\em Neural Networks}, 2(5):359--366, 1989.

\bibitem{huang2017densely}
Gao Huang, Zhuang Liu, Laurens Van Der~Maaten, and Kilian~Q Weinberger.
\newblock Densely connected convolutional networks.
\newblock In {\em CVPR}, pages 4700--4708, 2017.

\bibitem{jaynes1957information}
Edwin~T Jaynes.
\newblock Information theory and statistical mechanics.
\newblock {\em Physical Review}, 106(4):620, 1957.

\bibitem{kim2021co}
JangHyun Kim, Wonho Choo, Hosan Jeong, and Hyun~Oh Song.
\newblock Co-mixup: Saliency guided joint mixup with supermodular diversity.
\newblock In {\em ICLR}, 2021.

\bibitem{kim2020puzzle}
Jang-Hyun Kim, Wonho Choo, and Hyun~Oh Song.
\newblock Puzzle mix: Exploiting saliency and local statistics for optimal
  mixup.
\newblock In {\em ICML}, pages 5275--5285. PMLR, 2020.

\bibitem{kingma2015adam}
Diederik~P Kingma and Jimmy Ba.
\newblock Adam: A method for stochastic optimization.
\newblock In {\em ICLR}, 2015.

\bibitem{lin2014microsoft}
Tsung-Yi Lin, Michael Maire, Serge Belongie, James Hays, Pietro Perona, Deva
  Ramanan, Piotr Doll{\'a}r, and C~Lawrence Zitnick.
\newblock Microsoft coco: Common objects in context.
\newblock In {\em ECCV}, pages 740--755. Springer, 2014.

\bibitem{natarajan2013learning}
Nagarajan Natarajan, Inderjit~S Dhillon, Pradeep~K Ravikumar, and Ambuj Tewari.
\newblock Learning with noisy labels.
\newblock In {\em NIPS}, pages 1196--1204, 2013.

\bibitem{paszke2019pytorch}
Adam Paszke, Sam Gross, Francisco Massa, Adam Lerer, James Bradbury, Gregory
  Chanan, Trevor Killeen, Zeming Lin, Natalia Gimelshein, Luca Antiga, et~al.
\newblock Pytorch: An imperative style, high-performance deep learning library.
\newblock In {\em NIPS}, volume~32, 2019.

\bibitem{petit2018handling}
Olivier Petit, Nicolas Thome, Arnaud Charnoz, Alexandre Hostettler, and Luc
  Soler.
\newblock Handling missing annotations for semantic segmentation with deep
  convnets.
\newblock In {\em Deep Learning in Medical Image Analysis and Multimodal
  Learning for Clinical Decision Support}, pages 20--28. Springer, 2018.

\bibitem{qin2020resizemix}
Jie Qin, Jiemin Fang, Qian Zhang, Wenyu Liu, Xingang Wang, and Xinggang Wang.
\newblock Resizemix: Mixing data with preserved object information and true
  labels.
\newblock {\em arXiv preprint arXiv:2012.11101}, 2020.

\bibitem{rajpurkar2017chexnet}
Pranav Rajpurkar, Jeremy Irvin, Kaylie Zhu, Brandon Yang, Hershel Mehta, Tony
  Duan, Daisy Ding, Aarti Bagul, Curtis Langlotz, Katie Shpanskaya, et~al.
\newblock Chexnet: Radiologist-level pneumonia detection on chest x-rays with
  deep learning.
\newblock {\em arXiv preprint arXiv:1711.05225}, 2017.

\bibitem{rumelhart1986learning}
David~E Rumelhart, Geoffrey~E Hinton, and Ronald~J Williams.
\newblock Learning representations by back-propagating errors.
\newblock {\em Nature}, 323(6088):533--536, 1986.

\bibitem{shi2021marginal}
Gonglei Shi, Li Xiao, Yang Chen, and S~Kevin Zhou.
\newblock Marginal loss and exclusion loss for partially supervised multi-organ
  segmentation.
\newblock {\em Medical Image Analysis}, page 101979, 2021.

\bibitem{sohn2020fixmatch}
Kihyuk Sohn, David Berthelot, Nicholas Carlini, Zizhao Zhang, Han Zhang,
  Colin~A Raffel, Ekin~Dogus Cubuk, Alexey Kurakin, and Chun-Liang Li.
\newblock Fixmatch: Simplifying semi-supervised learning with consistency and
  confidence.
\newblock In {\em NIPS}, volume~33, pages 596--608, 2020.

\bibitem{uddin2021saliencymix}
AFM~Shahab Uddin, Mst~Sirazam Monira, Wheemyung Shin, TaeChoong Chung, and
  Sung-Ho Bae.
\newblock Saliencymix: A saliency guided data augmentation strategy for better
  regularization.
\newblock In {\em ICLR}, 2021.

\bibitem{vahdat2017toward}
Arash Vahdat.
\newblock Toward robustness against label noise in training deep discriminative
  neural networks.
\newblock In {\em NIPS}, volume~30, pages 5601--5610, 2017.

\bibitem{wang2017chestx}
Xiaosong Wang, Yifan Peng, Le Lu, Zhiyong Lu, Mohammadhadi Bagheri, and
  Ronald~M Summers.
\newblock Chestx-ray8: Hospital-scale chest x-ray database and benchmarks on
  weakly-supervised classification and localization of common thorax diseases.
\newblock In {\em CVPR}, pages 2097--2106, 2017.

\bibitem{xu2021partially}
Yanyu Xu, Xinxing Xu, Lei Jin, Shenghua Gao, Rick Siow~Mong Goh, Daniel~SW
  Ting, and Yong Liu.
\newblock Partially-supervised learning for vessel segmentation in ocular
  images.
\newblock In {\em MICCAI}, pages 271--281. Springer, 2021.

\bibitem{yun2019cutmix}
Sangdoo Yun, Dongyoon Han, Seong~Joon Oh, Sanghyuk Chun, Junsuk Choe, and
  Youngjoon Yoo.
\newblock Cutmix: Regularization strategy to train strong classifiers with
  localizable features.
\newblock In {\em ICCV}, pages 6023--6032, 2019.

\bibitem{zhang2018mixup}
Hongyi Zhang, Moustapha Cisse, Yann~N Dauphin, and David Lopez-Paz.
\newblock mixup: Beyond empirical risk minimization.
\newblock In {\em ICLR}, 2018.

\bibitem{zhang2021dodnet}
Jianpeng Zhang, Yutong Xie, Yong Xia, and Chunhua Shen.
\newblock Dodnet: Learning to segment multi-organ and tumors from multiple
  partially labeled datasets.
\newblock In {\em CVPR}, pages 1195--1204, 2021.

\bibitem{zhou2019prior}
Yuyin Zhou, Zhe Li, Song Bai, Chong Wang, Xinlei Chen, Mei Han, Elliot Fishman,
  and Alan~L Yuille.
\newblock Prior-aware neural network for partially-supervised multi-organ
  segmentation.
\newblock In {\em ICCV}, pages 10672--10681, 2019.

\bibitem{zhu2002learning}
Xiaojin Zhu and Zoubin Ghahramani.
\newblock Learning from labeled and unlabeled data with label propagation.
\newblock Technical Report CMU-CALD-02-107, Carnegie Mellon University, 2002.

\end{thebibliography}
}

\end{document}